\documentstyle[times,11pt,single-col,lingmacros]{article}

\input{psfig}
\input{psfig}
\begin{document}

\title{Punctuation in Quoted Speech}

\author{{\bf Christine Doran}  \thanks{I would like to thank Aravind
Joshi, Ted Briscoe, Ellen Prince, Beth Ann Hockey, B. Srinivas and
Jeff Reynar for their various contributions to this work. This
research has been partially supported by NSF grant NSF-STC SBR
8920230, ARPA grant N00014-94 and ARO grant DAAH04-94-G0426.} \\
\normalsize Department of Linguistics\\ \normalsize University of Pennsylvania \\
\normalsize Philadelphia, PA 19103 \\ \normalsize {\tt cdoran@linc.cis.upenn.edu}}
\date{}

\maketitle

\renewcommand{\baselinestretch}{1.0}

\begin{quote}
Quoted speech is often set off by punctuation marks, in particular
quotation marks.  Thus, it might seem that the quotation marks would
be extremely useful in identifying these structures in
texts. Unfortunately, the situation is not quite so clear. In this
work, I will argue that quotation marks are not adequate for either
identifying or constraining the syntax of quoted speech. More useful
information comes from the presence of a quoting verb, which is either
a verb of saying or a punctual verb, and the presence of other
punctuation marks, usually commas. Using a lexicalized grammar, we can
license most quoting clauses as text adjuncts. A distinction will be
made not between direct and indirect quoted speech, but rather between
adjunct and non-adjunct quoting clauses. 
\end{quote}

\parskip 1ex

\section{Motivation}
In looking at the ways punctuation can be used to help identify
particular structures in text, it might seem as if quotation marks
would be extremely useful. In particular, they might be useful in
distinguishing direct and indirect speech, which appear to have
radically different forms and functions. This would facilitate text
processing of such constructions, whether via full syntactic parsing
or some more superficial analysis, such as regular expression
matching. Unfortunately, the situation is not quite so clear. In
particular, the distinction between direct and indirect quoted speech
is very blurry.

The goal of the work described here is to untangle the syntax of
direct and indirect speech. The framework within which the present
work is couched is Lexicalized Tree Adjoining Grammar, which will be
briefly described in section \ref{TAG}. I will argue that the
direct/indirect split is not the correct one, and that the choice of
the verb and the other punctuation marks involved are more informative
than the quotation marks.

\section{What are the kinds of quoted speech?}

The first problem is to identify a class of constructions identifiable
as Quoted Speech. Punctuation-wise, we canonically expect a comma
after the quoting verb and quotation marks around the speech for
direct speech, and neither of these for indirect speech.\footnote{I am
leaving aside a possible third category, ``Free Indirect Speech''
which is argued to be an intermediary type, reflecting the sequence of
tense effects of indirect speech and the deictic use of direct
speech. For the features I am considering, it appears to pattern with
direct speech.} And indeed, there are clear cases of direct speech,
like (\ex{1}), and indirect speech (\ex{2}).

\enumsentence{A Lorillard spokeswoman said, ``This is an old
story. We're talking about years ago before anyone heard of asbestos
having any questionable properties.  There is no asbestos in our
products now.'' \hfill [wsj0003]}

\enumsentence{However, Mr. Dillow said he believes
that a reduction in raw material stockbuilding by industry could lead
to a sharp drop in imports. \hfill [wsj1500]}

\noindent
However, there are also cases which blur the distinction, such as
(\ex{1}) and (\ex{2}):

\enumsentence{Some bulk shipping rates have increased ``3\% to 4\% in
the past few months,'' said Salomon's Mr. Lloyd. \hfill [wsj1500]}

\enumsentence{And, they warn, any further drop in the government's
popularity could swiftly make this promise sound hollow. \hfill [wsj1500]}

\enumsentence{Republican Sen. William Cohen of Maine, the panel 's vice
chairman, said of the disclosure that ``a text torn out of context
is a pretext, and it is unfair for those in the White House who are
leaking to present the evidence in a selective fashion.'' \hfill [wsj1500]}

\noindent Example (\ex{-2}) is partly a direct quote (the object of
the verb) and partly indirect.  Example (\ex{-1}) has the usual
subject-verb-complement order (SVO), but has the subject and the verb
of saying separated from the speech by a comma. Example (\ex{0}) has
the syntax of an indirect quote, but uses quotation marks.

Furthermore, how are we to distinguish quoted material in the running
text from quoted speech proper? Examples (\ex{1})-(\ex{5}) show
several such variants. Text in scare quotes, terminology and other quoted
material included in running text are often only identifiable by their
enclosure in quotation marks, and they are not distinguished
syntactically from the surrounding material.

\enumsentence{...noted that the term ``teacher-employee'' (as
opposed to, e.g., ``maintenance employee'') was a not inapt
description.  \hfill [wsj1500]}

\enumsentence{Unable to persuade the manager to change his
decision, he went to a ``company court'' for a hearing. \hfill [wsj1500]}

\enumsentence{Mr. Nagrin has described four ``places'', each with
its scenery and people, added two ``diversions''... \hfill [Brown:cc09]}

\enumsentence{Types of loans SBA business loans are of two types: ``participation'' and ``direct'' \hfill [Brown:ch01]}

\begin{figure}[htb]
\centering
\begin{tabular}{cc}
{\psfig{figure=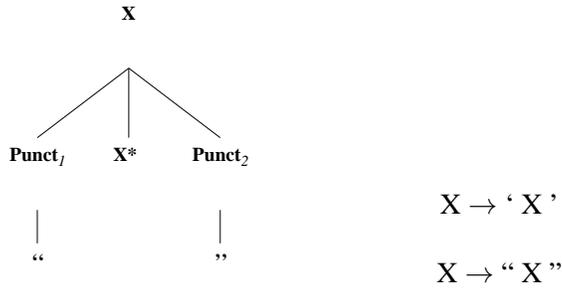,height=1.5in}} & \hspace{0.75in}
\vspace {-0.55in} X $\rightarrow$ `` X '' \\
 & \hspace{0.75in} X $\rightarrow$ ` X ' \\
\end{tabular}
\vspace{.5in}
\caption{The schematic LTAG tree and phrase-structure rule for
handling quotation marks, where X can be any node label. The tree is
lexicalized on both the opening and closing quotation marks, so we are
guaranteed to always get matching pairs of quotes.}
\label{quote-tree}
\end{figure}

From this data it seems clear that the quotation marks are not a
useful indicator of any particular construction. While text in
quotation marks is always a quotation of some sort, not all quotations
are enclosed in quotation marks.  Direct speech is simply a subset of
the more general class of (ostensibly) verbatim text. Semantically,
the material enclosed in quotation marks is a uniform class. The
quotation marks always have essentially the same semantics: they mark
what someone other than the author says/said/thinks/thought.  As with
scare quotes, the Other need not be identified explicitly. However,
the quotation marks themselves are not an indicator of the larger
syntactic context. Syntactically, we simply need a tree or a rule like
those in Figure \ref{quote-tree} to handle quotation marks.

But all is not lost --- I believe that the comma (or, less commonly,
the dash or colon) which appears in direct speech is actually the
important cue, along with the particular verb used in the quoting
clause.  In the remainder of the paper, I will argue that the relevant
distinction is between indirect quoted speech in the normal SVO order
and all other quoted speech, rather than between direct and indirect
speech.

\section{Lexicalized Tree Adjoining Grammar}
\label{TAG}

\begin{figure}[htb]
\centering
\mbox{}
\begin{tabular}{cccc}
{\psfig{figure=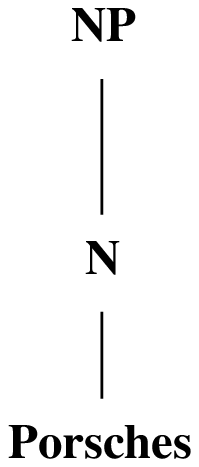,height=1.5in}} &
{\psfig{figure=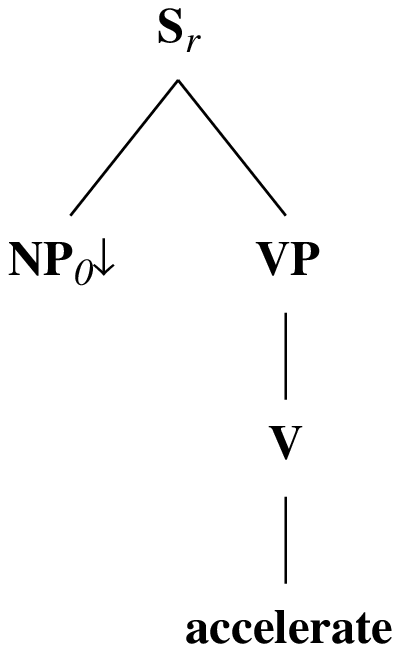,height=2in}} &
{\psfig{figure=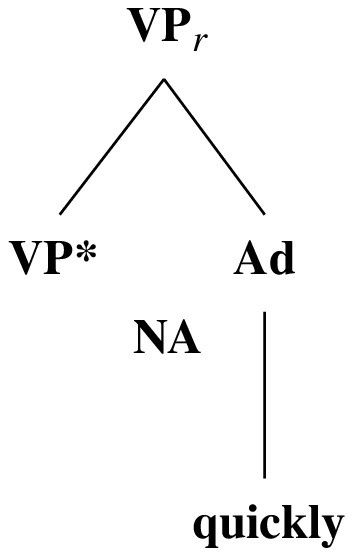,height=1.5in}} &
{\psfig{figure=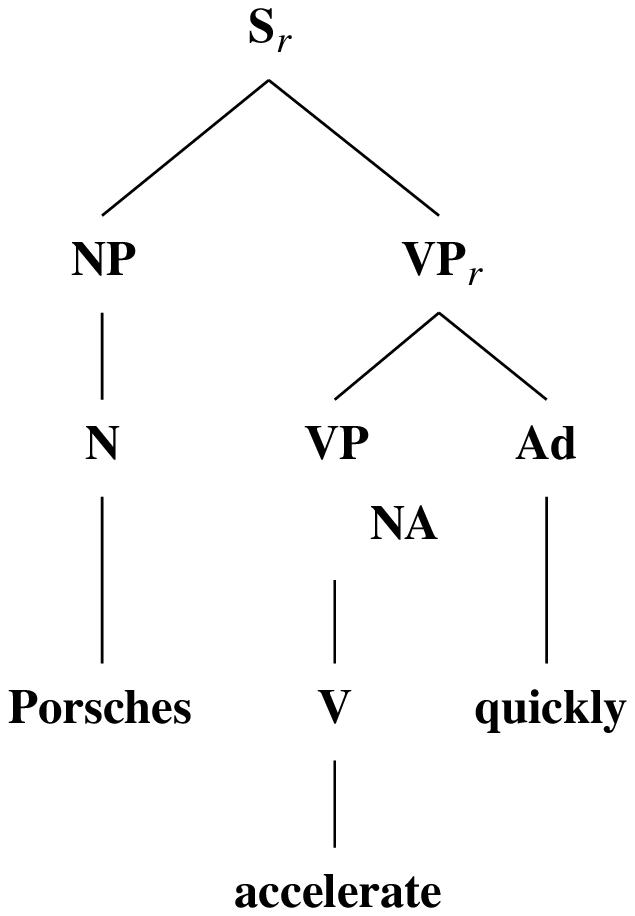,height=2.25in}} \\
(a) & (b) & (c) & (d) \\
\end{tabular}
\caption{Basic LTAG trees: (a) initial NP tree, (b) initial S tree,
(c) auxiliary adverb tree, and (d) S with NP substituted and adverb
adjoined. }
\label{TAG-basic}
\end{figure}

Lexicalized Tree Adjoining Grammar is a grammar formalism which has
evolved from Tree Adjunct Languages, introduced in \cite{joshi75}. The
basic units of any TAG grammar are {\em elementary trees}, of which
there are two types: {\em initial} and {\em auxiliary}. Two combining
operations are used: {\em substitution} and {\em adjunction}.  Initial
trees contain only argument positions, marked with $\downarrow$, where
other initial trees must be substituted.  Trees \ref{TAG-basic}(a) and
(b) are both initial trees, and (d) shows (a) substituted as the
subject of (b). Auxiliary trees can also have argument positions, but
they differ in having a distinguished leaf called the {\em foot}
(marked with $*$) which has the same label as the root. These trees
adjoin, or are spliced, into other trees. Tree \ref{TAG-basic}(c) is
an auxiliary adverb tree, and in (d) it has adjoined at the VP node.

Lexicalization \cite{schabes88} requires that each elementary tree in
the grammar be associated with at least one lexical {\em anchor}
(possibly more than one, for instance in handling idioms). This has
the effect of consolidating the lexicon and the grammar. The grammar
used here is fully lexicalized, and uses feature structures
\cite{vijay91}.

Lexicalized TAG has been shown to have many linguistically appealing
properties, including an extended domain of locality --- all of the
arguments of an anchor are localized within a single elementary
tree. Thus, both syntactic and semantic dependencies are expressed
locally.  For discussion of some linguistic issues, see
\cite{kj85,frank92}.  This provides a elegant framework for handling
clausal level information, since each simple clause (usually a verb
and its arguments) is a single tree. In contrast, a context-free
grammar will not have both the subject and the complements of a verb
in the same rule, making it harder to specify local constraints.

This work is included in a large English LTAG which has been developed
as part of the XTAG project. XTAG is a wide-coverage grammar which
includes a morphological analyzer, a part-of-speech tagger, a large
syntactic lexicon, and a parser.  For more details, see
\cite{tech-rept95}.

\section{Untangling quoted speech}

Direct and indirect quoted speech may or may not be enclosed in
quotation marks, but they are syntactically distinguished.  Typically,
indirect speech is shown as the complement to a verb of propositional
attitude, like {\it say} or {\it believe}, as in (\ex{1}). Direct
speech may also use the same syntax, as shown in (\ex{2}). Although it
is typically restricted to occuring with verbs of saying (\ex{3}),
this appears to be a pragmatic rather than a syntactic/lexical
constraint. In a context where it is possible to know what the speaker
is thinking, in particular in text with an omniscient narrator, this
construction is fine (\ex{4}). There are also differences in the point
of view (i.e. choice of first or third person pronouns, other
deictics) and in sequence of tense effects.

\enumsentence{After a few minutes he said (that) he couldn't use her
if she danced like that.}
\enumsentence{After a few minutes he said, ``I can't use you if
you dance like that.'' \hfill [cf09]}
\enumsentence{\#After a few minutes he believed/thought, ``I can't
use you if you dance like that.''} 
\enumsentence{Alice was beginning to get very tired of sitting by her
sister on the bank, and of having nothing to do: once or twice she had
peeped into the book her sister was reading, but it had no pictures or
conversations in it, `and what is the use of a book,' thought Alice,
`without pictures or conversation?' [First line of \underline{Alice's
Adventures in Wonderland}]}

However, direct speech has further options unavailable to indirect
speech. Direct speech may be introduced by punctual verbs like {\it
begin} and {\it continue}, as in (\ex{1}); these typically take
infinitival complements, so they cannot be
used with indirect speech (\ex{2}).

\enumsentence{A Birmingham newspaper printed in a column for children
an article entitled ``The Story of Guy Fawkes'', which began:``When you pile your ``guy'' on the bonfire tomorrow night...
\hfill [cd03]}

\enumsentence{$^*$A Birmingham newspaper printed an article which began
that when you pile your ``guy'' on the bonfire tomorrow night...}

Corpus data shows that direct speech is far less likely to occur with
a complementizer (although it can), and is more likely to have a comma
(dash, colon) before the complement clause. Direct speech usually has
quotation marks around the speech, but they are not required
(e.g. dialogues in works of fiction).
Both indirect and direct speech allow for multiple locations of the
quoting clause (the subject and the main verb) relative to the quoted
material: sentence initially, sentence finally and sentence
internally. This is the first issue upon which I will focus, as we
must first decide whether these variants are derivationally related to
the SVO order.

Finally, both types of speech can occur with intransitive or
transitive clausal complement verbs, as in examples (\ex{1}) and
(\ex{2}):

\enumsentence{Because of deteriorating hearing, she told colleagues
she feared she might not be able to teach much longer.\hfill
[wsj0044]}

\enumsentence{Richard Driscoll, vice chairman of Bank of New England,
told the Dow Jones Professional Investor Report, ``Certainly, there
are those outside the region who think of us prospectively as a good
partner.\hfill[wsj0067]}

\subsection{Position of the quoting clause}

In addition to preceding the speech, the subject and verb may follow
(\ex{1}) or be embedded in the speech (\ex{2}). The verbs which are
possible are the same for all orders of indirect and direct
speech. The subject and verb may be inverted in all of these positions
with the intransitive clausal complement verbs, as in (\ex{3}),
although one rarely finds inversion in the sentence initial position
in modern texts. Inversion of pronouns is also rare in modern texts ---
example (\ex{4}) is from Jane Austen's \underline{Persuasion}. No
complementizers are permitted with the embedded and sentence-final
orders. Finally, a preposition/subordinating conjunction is possible
before a quoting clause, as in (\ex{5}). {\it As} is the most commonly
used.

\enumsentence{``You can't do this to us'', Diane screamed.  ``We are
Americans''.        \hfill [Brown:cf09]}

\enumsentence{``Today 's action,'' Transportation Secretary Samuel Skinner
said, ``represents another milestone in the ongoing program
to promote vehicle occupant safety in light trucks and minivans
through its extension of passenger car standards.'' \hfill [wsj0064]}

\enumsentence{``The morbidity rate is a striking finding among those
of us who study asbestos-related diseases,'' said Dr. Talcott.
\hfill [wsj0003]}

\enumsentence{``That is the woman I want'', said he.  ``Something a
little inferior I shall of course put up with, but it must not be
much.  If I am a fool, I shall be a fool indeed, for I have thought on
the subject more than most men.''}

\enumsentence{But he, as I can now retort, was the man who could see
so short a distance ahead...   \hfill [Brown:cg70]}

The inversion is unusual in that it involves a main verb, and English
does not generally allow main verbs to invert. The syntactic details
are not crucial for the current purposes, but for a detailed
Minimalist account of quotative inversion, see Collins and Branigan
\shortcite{collins-branigan}. Their basic argument is that there is a null
operator in Spec/CP. The operator raises from the complement position
of the verb, where it leaves a co-indexed trace.  (They claim that it
can occasionally be lexicalized as {\it so} --- ``So Mary said.'') The
operator is bound by the quoted clause at a discourse level (similar
to PRO$_{arb}$).  Given the syntactic free choice between inverted and
non-inverted quoting verbs, there is clearly more to say about why one
form or the other is used. There appear to be interesting pragmatic
constraints on the inverted form, but I will not be discussing this
issue further here.

This positional variation raises interesting syntactic questions: are
the various orders derived from the sentence-initial order, with the
quoted clause always being an argument of the quoting verb? or are the
quoting clauses text adjuncts, like parentheticals, adjoining into
clauses at will? If the latter, do all of the orders behave alike?
\cite{emonds} argues that both the sentence-initial and sentence-final
orders are basic, and that the sentence-medial order is derived from
the latter. In that case, do the sentence-initial orders of both
direct and indirect speech have the same syntax, or do they diverge?
Let us consider each of the positions for the quoting clause in turn,
and see what they have to tell us about the larger syntactic picture.

\subsubsection{Sentence-internal order}

A movement analysis, where the speech is the complement of the quoting
verb, is impossible because it would require some sort of
``intraposition'' wherein the matrix clause moves \underline{into} the
quoting clause.  An alternative movement analysis which treats the
subject as topicalized is also impossible, since the quoting clause
occurs at any constituent boundary (modulo heaviness
effects). Examples (\ex{1}) and (\ex{2}) show a quoting clauses coming
between the verb and its complement. This is not typical of a
topicalization structure.

\enumsentence{``I rather resent'', she said, ``you speaking to those
groups in Portland as though just the move accomplished this.'' \hfill
[Brown:ca23]}

\enumsentence{Suppose, says Dr. Lyttleton, the proton has a slightly
greater charge than the electron so slight it is presently
immeasurable. \hfill [Brown:cc13]}

\begin{figure}[htb]
\centering
\mbox{}
\begin{tabular}{cc}
{\psfig{figure=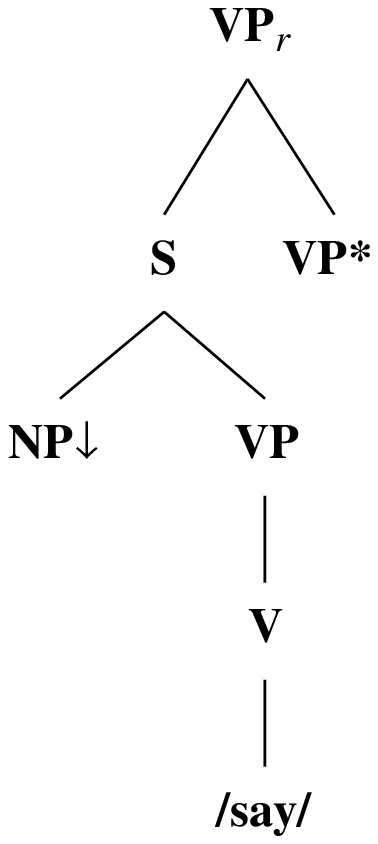,height=2.15in}} & \hspace{1.0in}
{\psfig{figure=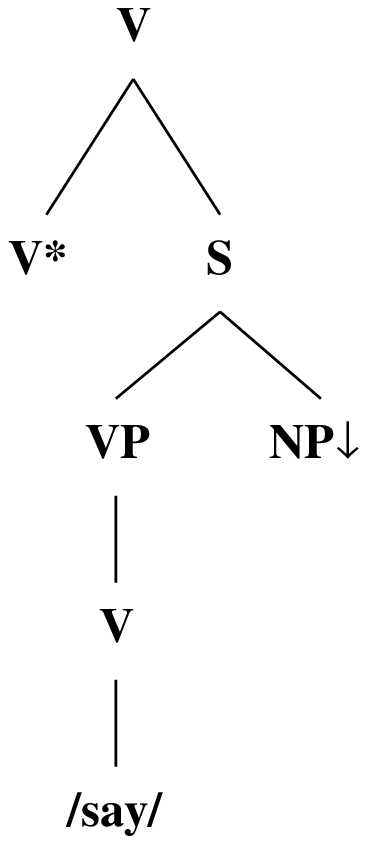,height=2.15in}} \\
(a) & (b) \\
\end{tabular}
\caption{The trees used for a non-inverted quoting clause: (a) pre-VP
e.g. {\it ``Today's action,'' Transportation Secretary Samuel Skinner
said, ``represents another...} and (b) post-V, e.g. {\it ``I rather
resent'', she said, ``you speaking...''} }
\label{paren-tree}
\end{figure}

Thus, we are led to an adjunct analysis of the quoting clause. LTAG is
very well-suited to such an analysis, because, as noted above, the
clause into which the quoting clause adjoins is in itself a complete
matrix sentence. Thus, there are no concerns about passing agreement
or other clausally local information ``across'' the parenthetical
quoting clause.  Sample LTAG trees for pre-VP and post-V quoting
clauses are shown in Figure \ref{paren-tree}.

Because the grammar is lexicalized, we can elegantly capture the
generalization that only verbs taking clausal complements can select
this structure. The LTAG lexicon groups clausal trees into {\em Tree
Families}, which contain all of the constructions allowed for a single
subcategorization frame (active, passive, wh- question, relative
clauses, etc.). These adjunct trees would simply be members of the
clausal complement tree families.\footnote{As text-adjuncts, they
would naturally have a different semantic interpretation from other
adjunction structures.} Furthermore, as I mentioned earlier, the LTAG
trees have a larger domain of locality than context-free grammars. In
this tree, the relationship between the quoting verb and the clause it
adjoins into is expressed in a single rule, allowing us to state
constraints imposed by the verb directly.

On this analysis, the quoted clause is not overtly a complement of the
quoting verb. However, each tree in a tree family is associated with a
type, which gives the number and type of arguments the verb
requires. The transitive family would have the type {\em NP x NP} and
the intransitive clausal complement family, the type {\em NP x
S}. When an argument is not overtly realized, as in an agentless
passive, this information is available to the semantic and discourse
modules. For the passive, we can look for the agent in the discourse
context, while for the quoting clause, the semantic component could
associate the matrix clause with the missing complement. If one wanted
a more explicit connection, a null operator as in
\cite{collins-branigan} could be built into the adjunct tree.

There are a number of reasons to believe that this is the correct
analysis. In this construction, verbs appear to lose many of their
selectional restrictions. As noted above, punctual verbs usually
select for infinitival complements, yet they can be embedded in tensed
quoted clauses.  Verbs also lose their selectional restrictions as to
{\em wh} features, with verbs like {\it insist} embedding in questions
as in (\ex{1}).

\enumsentence{Who, Mary insisted, has ever seen a purple elephant?}

Furthermore, the embedded quoting clauses are frequently synonymous
with other kinds of parentheticals: {\it John, I presume/presumably/it
appears, bought a new car}. Like other parentheticals, quoting clauses
are also set off with ``comma intonation'' in speech.
\cite{schmidt95} finds that there is a significant pitch range
restriction across parenthetical types, but he does not give any
examples of direct quotation. While we should be cautious about
drawing analogies between prosody and punctuation, if quoted clauses
were shown to have a similarly restricted pitch range, this would be
further evidence for the similarity of the constructions.

In his discussion of parentheticals as discontinuous constituents,
McCawley \shortcite{mccawley82} argues that the parenthetical does not
behave as part of the constituent that contains it. The ellipsis tests
he use to support his argument suggest that the quoting clauses behave
similarly. 

\enumsentence{John, Mary said, bought a house, and Sue did too $=$ Sue
bought a house {\sc or} Sue said John bought a house $\neq$ Mary said
Sue bought a house}

\noindent
In (\ex{0}), the antecedent for the ellipsis is {\it said} or {\it
bought}, but not {\it said..bought}. This is what would be predicted
if the complete sentence is not a constituent available as an
antecedent.

\subsubsection{Sentence-final position}

In this order, it is certainly more plausible that the complement
clause is fronted. However, Emonds gives some compelling examples
against a derivational relation.

\enumsentence{John hasn't completed his book, I don't think. \hfill [Emonds' II.91]} 
\enumsentence{John hasn't completed his book, I think.}
\enumsentence{I don't think John hasn't completed his book.}
\enumsentence{I think John hasn't completed his book.}   

Sentences (\ex{-3}) and (\ex{-2}) are synonymous for speakers who
accept both variants, i.e. the negation in the quoting clause has no
effect. However, in the sentences these would have to be derived from,
(\ex{-1}) or (\ex{0}), the presence or absence of the matrix or
negation does change the meaning of the sentence.

\begin{figure}[ht]
\centering 
\mbox{}
{\psfig{figure=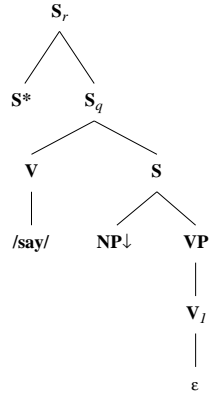,height=2.15in}}
\caption{The tree used for an inverted, post-S quoting clause,
e.g. {\it  `Come, let's try the first figure!' said the Mock Turtle to the
Gryphon.} \hfill [Carroll:AAIW]}
\label{s-final-tree}
\end{figure}

Additionally the sentence-final order for quoting clauses shares the
features of embedded quoting clauses discussed in the previous
section, and we again are led to decide against a movement analysis
and for an adjunct analysis. Figure \ref{s-final-tree} shows the
relevant tree.

\subsubsection{Sentence-initial position}

\begin{figure}
\centering
\mbox{}
{\psfig{figure=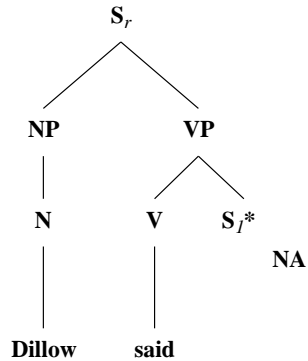,height=2.0in}}
\caption{The basic LTAG tree for clausal complements.}
\label{reg-scomp}
\end{figure}

Finally, we come to the most difficult case --- the quoting clause in
sentence initial position. As in the other cases, direct and indirect
speech are identical in the left to right order of constituents. In
the previous two sections, direct and indirect speech patterned
together, as parenthetical clauses. However, the question here is
whether they will continue to pattern together.  The LTAG analysis for
normal clausal complement structures is shown in Figure
\ref{reg-scomp}. The tree adjoins at the root of the complement clause
tree for indicative clausal complements and below the extracted
element in extracted clause. This analysis gives an elegant treatment
of long-distance extraction (see \cite{kj85}).

We could simply allow the additional punctuation to adjoin to this
tree for sentences like (\ex{1}). 

\enumsentence{Alice replied very readily: `but that's because it stays
the same year for such a long time together.' \hfill [Carroll:AAIW]}

However, if we look more closely at the two kinds of speech, we find
several differences. For one, direct speech requires that questions be
inverted, (\ex{1}) and (\ex{2}), while normal clausal complements
cannot be inverted, (\ex{3}) and (\ex{4}).

\enumsentence{Alice asked,  `Has anyone seen the Cheshire Cat?' }
\enumsentence{$^*$Alice asked (whether), `Anyone had seen the Cheshire Cat?' }
\enumsentence{$^*$Alice asked (whether) has anyone seen the Cheshire
Cat. }
\enumsentence{Alice asked whether anyone had seen the Cheshire
Cat. }

Secondly, you cannot get embedding in the quoting clause of direct
speech, whereas you can have (in principle) unbounded embedding in
clausal complements:

\enumsentence{$^*$The queen said the White Rabbit whispered Alice
asked, `Has anyone seen the Cheshire Cat?' } 
\enumsentence{The queen said the White Rabbit whispered...that Alice
asked whether anyone had seen the Cheshire Cat.}

These differences suggest that Emonds was correct in concluding that
the quoted clause in the parenthetical type of quoted speech is a
matrix sentence.
Thus, there are two kinds of derivations possible for sentence initial
quoting clauses. If there is no punctuation other than quotation marks
after the quoting verb, we use the tree in Figure \ref{reg-scomp}. If
there is punctuation, the LTAG tree would be identical to Figure
\ref{s-final-tree}, but would have the foot node on the right. This
will mean giving sentences like (\ex{1}) the same analysis as
indirect speech, i.e. the non-parenthetical analysis.

\enumsentence{Gemina said in a statement that "it reserves the right
to take any action to protect its rights as a member of the
syndicate." \hfill [wsj1371]}

\section{What about quote transposition?}

As the alert reader will have noticed, the example trees given thus
far do not contain any punctuation marks. The quotation marks would be
handled as shown in Figure \ref{quote-tree} above, simply adjoining
onto the quoted constituent.  Given that we have made the
relevant distinction parenthetical quoting clauses
vs. non-parenthetical quoting clauses, it is clear that the comma
(dash, colon) separating the quoting clause from the quote is crucial
to the construction.

As Nunberg \shortcite{nunberg90} discusses at some length, American
English and British English differ in how they treat closing
punctuation next to a closing quote. In American English, commas and
periods are transposed with closing quotation marks (e.g. {\bf .''}),
while in British English the comma or period remains outside of the
quote (e.g. {\bf ''.}). The Brown corpus (exclusively American texts)
shows this distinction to be unhelpful: there are only 39 commas and
28 periods inside of quotation marks (both single and double), but
1823 commas and 1023 periods outside. The so-called British system is
massively predominant. This may be the result of post-processing on
the corpus, since analysis of 2.5 million words of Wall Street Journal
data turns up only 15 commas and 15 periods in the British system.
Sampson \shortcite{sampson93} also cites an example from the LOB corpus of
British English which uses the American system. In any event, it is
clear that if one is dealing with naturally occurring data, one is
likely to encounter both systems in varying proportions.

So, how is one to (a) require the punctuation mark to be present and
(b) allow it to appear in either of two locations? 

\begin{figure}[htb]
\centering
\mbox{}
{\psfig{figure=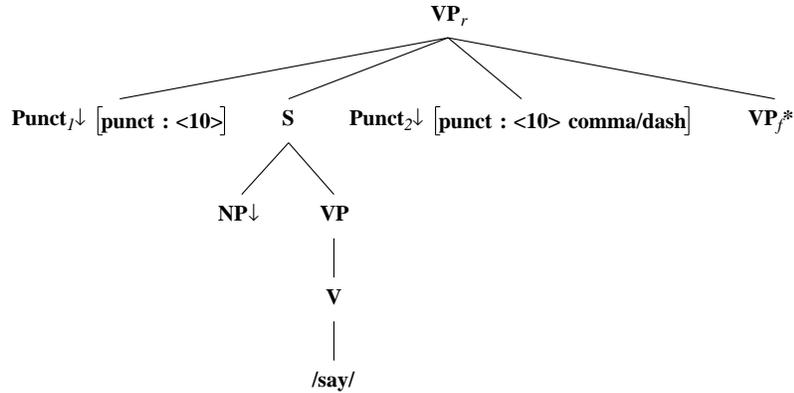,height=2.15in}}
\caption{Tree for embedded quoting clause, with
punctuation argument positions.}
\label{w/punct-tree}
\end{figure}

\begin{figure}[tb]
\centering
\mbox{}
{\psfig{figure=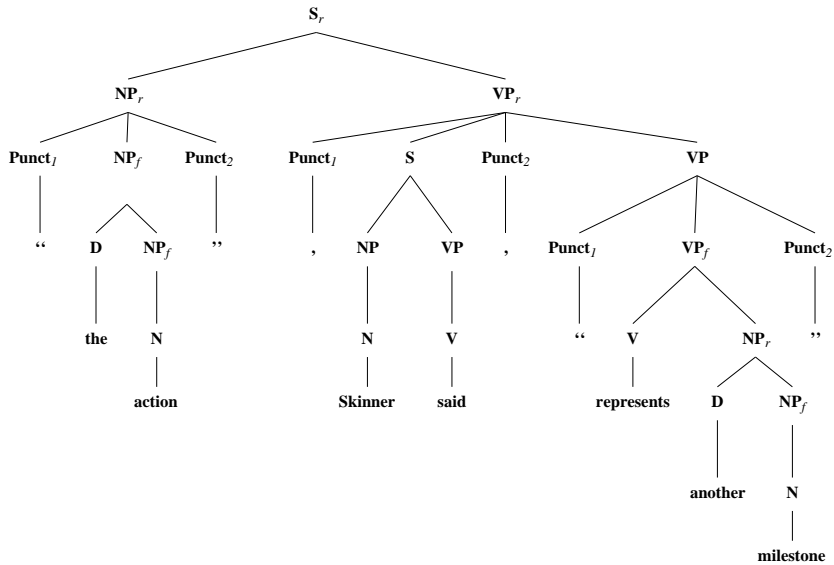,height=3.0in}}
\caption{Parsed sentence with embedded quoting clause and quotation
marks, British order.}
\label{w/embed-tree}
\end{figure}

Assuming that the quoting clause is like a parenthetical, the commas
around it are Nunberg's ``delimiting'' punctuation marks. However, I
will follow Briscoe \shortcite{briscoe} in allowing for both balanced
and unbalanced variants.  This is easily captured in the LTAG
treatment, since each position for the quoting clause has its own
tree. This also allows us to license a colon in the sentence initial
order, but not in either of the other orders, but let us leave aside
the colon for the moment. The tree for post-subject quoting clauses
is shown in Figure \ref{w/punct-tree}; the trees for pre-S and post-S
clauses would be similar, but would have only one {\sc Punct} node.
If the quoted constituent is an S, we can get the two orders of
quotation marks and commas by allowing the quotes to adjoin either
above or below the comma. Unfortunately, the quoted constituent may be
of any type. Using tree \ref{w/punct-tree}(b), the first pair of
quotation marks is around the subject NP. With this tree, we will only
get the British order, Figure (\ref{w/embed-tree}).

An alternative would be to allow the comma to adjoin above the S for
the British order, or above the leftmost quoted constituent for the
American order. However, since we have determined that the punctuation
mark is required in the tree, it is more elegant to have it as an
argument. 

A second alternative would be to treat quote inversion as something
like clitic-climbing by the comma. This would allow us to use the same
tree for both orders, but the American order would use a
multi-component tree set. Briefly stated, a multi-component set allows
one to force a set of trees to act as single tree --- if one tree in the
set is used in a derivation, all of the trees must be used.  The two
components of this set would be a tree anchored by the trace, which would
substitute into the argument position, and a tree anchored by the
comma, which would adjoin to the closing quote. The same
multi-component set would be selected by both the comma and the
period, but not by the dash, exclamation point or question mark, as
they do not undergo quote inversion.\footnote{In fact, dashes quite
rarely set off quotative clauses and I suspect may only do so in the
clause internal position. It would be straightforward to capture this
with the features in the quoting clause LTAG trees.}

The simplest solution is to do some ``normalization'' in tokenizing
the data, in this case into the British form. This is the option which
we are currently pursuing. It might also be possible to address some
of the point absorption issues in the tokenization step.

\subsection{How to treat the colon}

With sentence-initial quoting clauses, a colon can sometimes follow
the verb of saying. Given that the colon is not typically a delimiting
punctuation mark, and that it does not participate in quote
transposition, we might well want to group constructions like (\ex{1})
with non-parenthetical quoted speech. Recall that the colon is
possible only in the sentence-initial order. Also,
complementizers are more freely permitted here. It remains to be seen
whether this construction takes inverted complements --- if it does not,
then it clearly ought to be classed with the non-parenthetical quoting
clauses.

\enumsentence{Indicating the way in which he has turned his back on
his 1910   philosophy , Mr. Reama said: ``A Socialist is a person
who believes in dividing everything he does not own''. \hfill [Brown:ca05]}

\noindent
Instead of having the {\em colon+} feature on tree
\ref{w/punct-tree}(a), we would have it on the foot S node of Tree
\ref{reg-scomp}.

\section{Quote Alternation}

American English requires that nested quotation marks alternate
between single and double marks, with double-quotes on the outermost
pair. (In British English, the outermost quotes are single.) This is
handled in the TAG account by a {\sc contains} feature, which is also
used to block self-embedding of other text-adjuncts, such as
appositives (cf. discussion in \cite{nunberg90}). The feature has the
value {\sc dquote+} at the root of the tree anchored by double
quotation marks (shown in Fig. \ref{quote-tree}), to indicate that the
subtree contains double quotes, and the value {\sc dquote--} on the
foot node, to block the tree from adjoining to any subtree which
already contains double quotes. The same feature is used with the
value {\sc squote+/--} for single quotes.  Note that since the grammar
is lexicalized (here, on the punctuation marks themselves) the
features come from different instantiations of a single tree (i.e. we
do not need separate trees for each type of quotation mark). Other
trees in the grammar are simply transparent to the {\sc contains}
feature, passing up its value in the relevant contexts. The quote
trees themselves are opaque to all other values of {\sc contains}, so
that for instance, while colon-expansions cannot usually be embedded,
they can be embedded if the inner expansion is inside of
quotation marks. 

\section{Conclusion}
I have shown that quotation marks are not adequate for either
identifying or constraining the syntax of quoted speech. More useful
information comes from the presence of a quoting verb, which is either
a verb of saying or a punctual verb, and the presence of other
punctuation marks, usually commas. Using a lexicalized grammar, we can
license most quoting clauses as text adjuncts, anchored by the
appropriate subset of verbs, and selecting the relevant punctuation
marks as arguments. Quoting clauses which are sentence-initial and are
not separated from the quote by a comma or dash are treated as normal
clausal complement verbs, with the quoted material as the internal
argument of the verb.

I have also shown that lexicalization and features as utilized by
LTAGs allow us to elegantly capture the distribution of both the verbs
and the punctuation marks in the relevant constructions.

\newpage
\parskip 0ex


\end{document}